\documentclass{article}

    \usepackage[preprint]{neurips_2025}

\usepackage[utf8]{inputenc} % allow utf-8 input
\usepackage[T1]{fontenc}    % use 8-bit T1 fonts
\usepackage{hyperref}       % hyperlinks
\usepackage{url}            % simple URL typesetting
\usepackage{booktabs}       % professional-quality tables
\usepackage{amsfonts}       % blackboard math symbols
\usepackage{nicefrac}       % compact symbols for 1/2, etc.
\usepackage{microtype}      % microtypography
\usepackage{xcolor}         % colors
\usepackage{xspace}
\usepackage{graphicx}

\newcommand{\xhdr}[1]{\vspace{0em}\noindent{{\bf #1.}}}
\newcommand{\ie}{\textit{i.e., \xspace}}
\newcommand{\eg}{\textit{e.g., \xspace}}

\usepackage{colortbl}

\definecolor{Gray}{gray}{0.9}
\definecolor{LightCyan}{rgb}{0.88,1,1}
\definecolor{darkgreen}{rgb}{0.1,0.7,0.1}
\newcolumntype{a}{>{\columncolor{Gray}}c}
\newcolumntype{b}{>{\columncolor{white}}c}

\usepackage{wrapfig}
\usepackage[most]{tcolorbox}  % advanced tcolorbox with skins

\tcbset{
  mybox/.style={
    colback=blue!5!white,
    colframe=blue!50!black,
    fonttitle=\small\bfseries,
    fontupper=\footnotesize,
    title=Multimodal Explanation,
    boxrule=0.5pt,
    arc=4pt,
    outer arc=4pt,
    left=2pt,
    right=2pt,
    top=2pt,
    bottom=2pt,
    coltitle=white,
  }
}

\title{Rethinking Explainability in the Era of Multimodal AI}

\author{%
  Chirag~Agarwal\\
  University of Virginia\\
  \texttt{chiragagarwal@virginia.edu} \\
}

\begin{document}

\maketitle

\begin{abstract}
    \looseness=-1 While multimodal AI systems (models jointly trained on heterogeneous data types such as text, time series, graphs, and images) have become ubiquitous and achieved remarkable performance across high-stakes applications, transparent and accurate explanation algorithms are crucial for their safe deployment and ensure user trust. However, most existing explainability techniques remain unimodal, generating modality-specific feature‐attribution, concepts, or circuit traces in isolation and thus failing to capture cross‐modal interactions. \textit{This paper argues that such unimodal explanations systematically misrepresent and fail to capture the cross-modal influence that drives multimodal model decisions, and the community should stop relying on them for interpreting multimodal models}. To support our position, we outline key principles for multimodal explanations grounded in modality: Granger-style modality influence (controlled ablations to quantify how removing one modality changes the explanation for another), Synergistic faithfulness (explanations capture the model's predictive power when modalities are combined), and Unified stability (explanations remain consistent under small, cross-modal perturbations). This targeted shift to multimodal explanations will help the community uncover hidden shortcuts, mitigate modality bias, improve model reliability, and enhance safety in high‐stakes settings where incomplete explanations can have serious consequences.  
\end{abstract}

\section{Introduction}
\label{sec:intro}
\looseness=-1 Recent advances in AI systems increasingly integrate multimodal data -- vision and language, audio and language, graph and language, or videos, audios, and language -- to achieve state-of-the-art performance~\citep{team2023gemini,hurst2024gpt,llama4} in complex multimodal tasks like visual-question answering~\citep{zhang2025vlm}, multilingual automatic speech recognition~\citep{conneau2023fleurs}, video question-answering~\citep{patraucean2023perception}. For instance, multimodal clinical models combining imaging, lab results, and text outperform unimodal models by 6–30\% margins~\citep{pahud2024orchestrating}. With the increasing development of multimodal models, it becomes imperative that their predictions are readily explainable to relevant stakeholders. However, \textbf{the field of explainable AI (XAI) has not kept pace with the multimodal surge} and most existing explainability techniques remain \textit{unimodal}, designed for models with a single input type -- explaining an image classifier with a heatmap~\citep{ribeiro2016should,fong2017interpretable,sundararajan2017axiomatic,selvaraju2020grad,agarwal2020explaining}, identifying circuit traces in large language models (LLMs)~\citep{ameisen2025circuit}, visualizing attention weights~\citep{galassi2020attention,danilevsky-etal-2020-survey}, or node- and edge-weights for graph models~\citep{ying2019gnnexplainer,agarwal2022probing,agarwal2023evaluating}. To faithfully explain a multimodal model, we must develop modality-aware techniques that reveal \textit{how multiple modalities interact} to produce their predictions and satisfy certain multimodal properties.

We argue that the AI community must \textbf{stop relying on unimodal explainability techniques in a multimodal world} as they yield misleading/incomplete explanations, missing the core aspect of multimodal model behavior. For instance, in multimodal sentiment analysis, a model may draw on both spoken audio and transcript text; an unimodal explanation could indicate that positive words in the text led to a positive sentiment prediction, missing that the sarcastic tone in the audio drove the model's negative classification. 

\looseness=-1\xhdr{Present work} We outline \textit{why unimodal XAI fails for multimodal systems} by examining model properties that unimodal explanations cannot capture. We illustrate these failures with case studies across different multimodal models (see Figs.~\ref{fig:position-xai}-\ref{fig:logit}), and provide key desiderata for multimodal explanations, \textit{viz.} i) Granger-style modality influence, \ie measuring how ablating one modality's explanation reshapes another's explanation, ii) Synergistic faithfulness, \ie matching each modality's true contribution to the model's output, and iii) Unified stability, \ie robustness under modality-specifc imperceptible perturbations, highlighting the urgent need for \textit{multimodal-aware explainability frameworks} -- a necessary step to keep our understanding of AI systems on pace with their growing complexity.

\textit{\textbf{Overall, in this work, we argue that relying on unimodal explainability for multimodal models is fundamentally flawed and demands a paradigm shift, revisiting the design of explanation algorithms, redefining evaluation protocols, and embracing truly multimodal evaluation criteria.}}

\section{Why Unimodal Explanations Fall Short for Multimodal Models?}
\label{sec:why}
\looseness=-1 Unimodal explanation techniques assume a single input source drives the model's decisions and thus generate individual explanations for each modality. Common approaches include saliency maps highlighting important pixels in an image~\citep{ribeiro2016should,fong2017interpretable,sundararajan2017axiomatic,selvaraju2020grad,agarwal2020explaining}, developing concept bottleneck layers for implicit interpretability~\citep{koh2020concept}, attention-weight visualization~\citep{galassi2020attention,danilevsky-etal-2020-survey}, node and edge weights for graph machine learning models~\citep{ying2019gnnexplainer,agarwal2022probing,agarwal2023evaluating}, and several other model-agnostic feature attribution methods that assign importance values to input features. These methods fall short in explaining multimodal models as they fail to capture several key properties of multimodal model behavior. We discuss these limitations below and why ignoring them leaves interpretations of multimodal systems fundamentally inaccurate.

\subsection{Unimodal Explanations Fail to Capture the Multimodal Attribution Gap}
\begin{figure}[t]
    \centering
    \includegraphics[width=\textwidth]{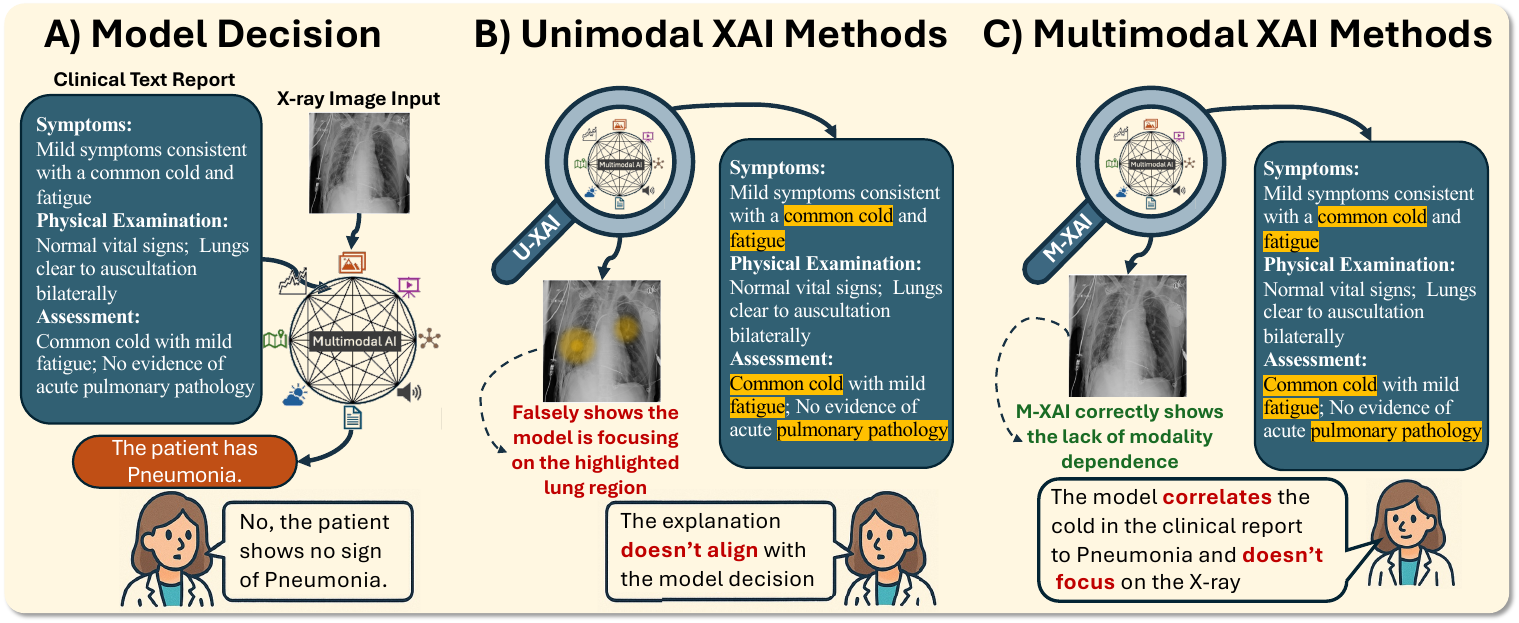}
    \caption{\looseness=-1\xhdr{Unimodal XAI (U-XAI) vs. Multimodal XAI (M-XAI) Attributions in Clinical Decision Support} \textbf{(A)} A multimodal diagnostic model that processes both a patient's clinical text report (symptoms, physical exam) and their chest X-ray to conclude, ``\textit{The patient has Pneumonia,}'' but a clinician, examining the case, notes no radiographic evidence of pneumonia. \textbf{(B)} Applying unimodal explainability method separately to each modality yields two disjoint attribution maps: a heatmap over the X-ray highlighting lung regions and a word-level highlight in the text report (\eg \textit{common cold}, \textit{fatigue}, \textit{pulmonary pathology}). Because these unimodal maps are generated in isolation, they \textcolor{red}{\textbf{falsely}} imply that the model relied on features from both modalities. 
    \textbf{(C)} In contrast, a multimodal XAI method produces an integrated explanation that reveals true modality dependence: it shows that the model's decision \textbf{correlates strongly with the phrase ``\textit{common cold}''} in the clinical report. By jointly analyzing both modalities, the multimodal explanation \textcolor{darkgreen}{\textbf{correctly}} identifies that the model ignored the image and ``\textit{didn't focus on the X-ray,}'' giving the clinician the insight needed to diagnose the shortcut behavior.}
    \vspace{-0.1in}
    \label{fig:position-xai}
\end{figure}

\looseness=-1 Most of the aforementioned explanation methods have been useful for explaining unimodal models in isolation~\citep{agarwal2022openxai}. For instance, Grad-CAM heatmaps~\citep{selvaraju2020grad} can pinpoint which part of an image influenced an object recognition result, and attention-based explanations can indicate which words a language model focused on for a prediction, examining the relationship between a model's prediction and the features of one modality at a time. However, for a multimodal model with inputs, say, from image $I$ and text $T$, a unimodal approach might produce an importance map over pixels or a highlight of words – essentially treating the image and text separately. Such an approach treats explainability for multimodal models as ``\textbf{explainability of parts}'' rather than explaining the whole (see Figure~\ref{fig:position-xai}), missing the fundamental fact that multimodal models are not merely the sum of two independent parts and a multi-modal model's behavior cannot be fully explained by dissecting each modality's influence in a silo. Unimodal explanations, by design, have no mechanism to represent such dependencies, and this limitation has practical consequences. This is further corroborated in~\citet{neo2024towards}, where the authors identify the most important tokens for the model's decision using Integrated Gradients~\citep{sundararajan2017axiomatic} and perform a token ablation experiment to impair the model's ability to perform object identification. Their results show that, for a significant number of ablated tokens, object token ablation consistently results in larger performance decreases across different visual-language models as compared to the gradient-based baseline, highlighting the ineffectiveness of gradient-based explanations in a multimodal setting.

\looseness=-1 Figure~\ref{fig:position-xai} illustrates the fundamental shortcomings of applying unimodal attribution methods to multimodal diagnostic models. Consider a clinical AI system that takes a patient's chest X-ray and accompanying textual notes to predict whether a patient has Pneumonia. A standard unimodal explanation pipeline would produce a heatmap over pulmonary opacities in the X-ray and a separate highlight of keywords, such as \textit{cough}, in the clinical report. On the surface, these attributions appear \textit{plausible}~\citep{agarwal2024faithfulness}—clinicians indeed associate lung haziness and cough with pneumonia—but they tell us nothing about the model's true inference path. Internally, the model may correctly fuse radiographic indicators with reported symptoms or exploit a spurious correlation by relying solely on the word ``cough,'' which frequently co-occurs with Pneumonia labels in the training data. Neither unimodal map can distinguish these possibilities: each is plausible in isolation, yet collectively insufficient to expose whether the model performed genuine multimodal reasoning or learned a text-only shortcut. This example highlights the need for multimodal explanation frameworks that disentangle and evaluate cross-modal contributions, rather than presenting fragmented, potentially misleading, and unimodal attributions.

\looseness=-1 To this end, some seemingly multimodal explanation approaches essentially develop specific architectures or are just collections of unimodal analyses~\citep{kim2017visual,park2018multimodal,wang2021interpretability,jung2021towards,sun2022explain,lyu2022dime,rodis2024multimodal,sun2024review}. For instance, one might apply a standard image attribution method and a separate text attribution method and present both results side by side. While it's better than nothing, it fails to reveal whether the model's decision process is more than the independent sum of its parts. Recent studies underscore that merging multiple data types and deriving a meaningful overall explanation is challenging~\citep{pahud2024orchestrating}. Only a multimodal explanation that faithfully (reflecting the true behavior of the model) links the symptom description with specific visual evidence (\eg ``\textit{cough in the note combined with right-lung opacity contributed jointly to the diagnosis}'') could reveal whether the model performed genuine multimodal reasoning. In fact, without such cross-modal insight, clinicians might falsely assume that the model validated the diagnosis radiologically, when it may have ignored the image altogether. This failure mode is not speculative: studies have shown that some radiology NLP systems predict disease from textual templates alone, without meaningful imaging assessment, especially when trained on weakly supervised datasets~\citep{casey2021systematic}. Unimodal explanations are ill-equipped to expose this failure mode. They 
tend to over-credit each modality's role when used in isolation, because they cannot show when one modality's input rendered the other modality effectively irrelevant in the model's reasoning.

\subsection{Misaligned Attention Weights Fail to Capture Cross-Modal Influence}
% Misaligned attention weights provide misleading explanations and fail to capture cross-modal influence.}
\looseness=-1 Attention weights have been widely used as a proxy for explanation, \eg looking at image attention or textual attention scores~\citep{nam2017dual,anderson2018bottom,xie2019visual,vijayaraghavan2021interpretable}, by model developers and stakeholders. Following this, many works use attention to understand multimodal models as they use attention mechanisms to align information (\eg how an image submodule attends to relevant words, and vice versa), which can be dangerously misleading~\citep{seth2025systematicevaluationhallucinationslargevision}. Attention reflects alignment, not causal influence. Just because a model attends to a feature does not mean that feature caused the decision. For instance, in a vision-language model, an image region with high attention might appear prominent, yet the model's prediction might remain unchanged if that region were masked~\citep{kang2025see}. Attention shows where signals are routed, not whether they were decisive. This is corroborated by recent works in language models that demonstrate that attention distributions can be re-weighted or manipulated without changing model predictions~\citep{jain2019attention,wiegreffe-pinter-2019-attention}. This non-robustness undermines the assumption that attention weights faithfully reflect what the model relies on. This is one of the key failures, where attention-based explanations are plausible without being correct~\citep{agarwal2024faithfulness}, \ie they generate clean heatmaps or token highlights that look explanatory, leading users to trust a model's process. But without a causal basis, \textbf{this is often an illusion}, especially if the model exploits dataset biases or shortcuts. 

To this end, multimodal reasoning often involves interdependencies between modalities, where visualizing modality-specific attention or cross-attention fails to reveal this inter-modal chain of influence. A reliable explanation must show how a cue in one modality alters processing in the other, something that attention alone cannot surface, as they fail to capture modality-level attribution. Attention maps do not provide a way to measure how much each modality (image vs. text or audio vs. text) contributed overall to the model's decision. In cases where one modality dominates, attention-based views might still highlight both inputs, obscuring the unimodal bias. Together, these issues make clear that attention visualizations, while intuitive, are not trustworthy explanations. In multimodal contexts, this is magnified, where attention can hide critical failures of reasoning, such as modality over-reliance or broken cross-modal alignment, and obscure spurious text-driven behavior.

\subsection{Mechanistic Interpretability Tools do not Scale to Multimodal Models}
\begin{figure}
    \centering
    \includegraphics[width=\textwidth]{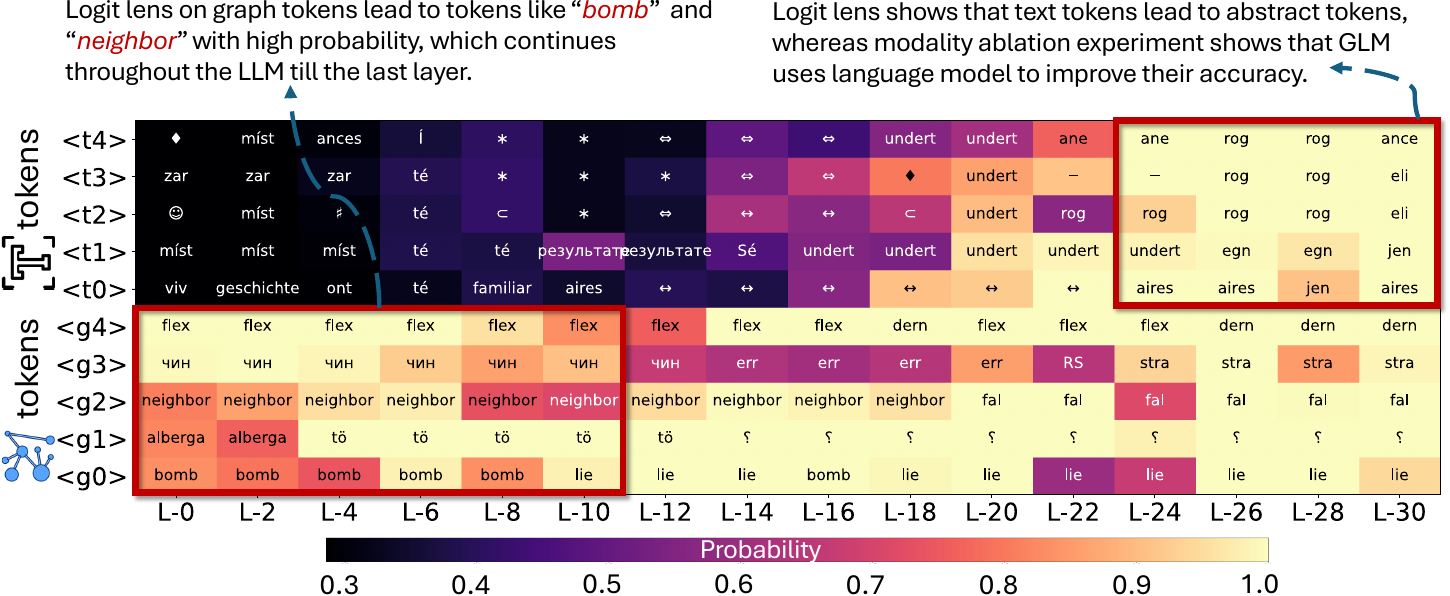}
    \caption{\looseness=-1\xhdr{Interpreting Graph Language Models using Logit Lens~\citep{interpre70}} The ``\textit{logit lens}'' provides a simple interpretability lens, designed to analyze token predictions across layers. Here, we use it to analyze a graph language multimodal model (GLM), which projects graph embeddings into the LLM's embedding space as specialized tokens ($g_i$) and is typically trained for node classification and link prediction (following~\citet{wang2024llms}, we trained the GLM on the Cora dataset~\citep{T8/HUIG48_2017} for node classification, where the correct label for this graph-language input is ``\textit{theory}''). Intuitively, the graph inputs, functioning as soft prompts, lack correspondence to the model's text token vocabulary, precluding interpretation via standard text-based analysis. Our analysis confirms that the logit lens, which was initially proposed for LLMs, \textbf{offers limited utility for GLMs}, and in some instances, can lead to abstract token predictions for text, as observed in prior works~\citep{neo2024towards}. We observe that graph token probabilities and vocabulary predictions remain static across LLM layers (here, we use Phi-3.5~\citep{abdin2024phi}; we visualized half of the layers due to space constraints and share more details in Appendix~\ref{app:experiment}), unlike the dynamic evolution of text token probabilities. The fundamental difference in training objectives between LLMs (unimodal) and GLMs (multimodal), particularly the autoregressive training of LLMs, which is absent in typical GLM training, creates a disparity in how graph and text tokens are processed. This further invalidates the direct application of text-centric interpretability methods to the graph modality and demonstrates the failure of unimodal techniques to provide interpretable insights for multimodal models, due to their implicit assumptions about model structure and training objectives.}
    \vspace{-0.1in}
    \label{fig:logit}
\end{figure}

\looseness=-1 Most existing explanation methods operate at the level of input-output, \ie highlighting input features that contribute to a model's prediction. While useful, such methods offer no visibility into the internal dynamics of the model. Because multimodal models learn to associate and transfer information between modalities, a key requirement for explainability is transparency of information flow inside the model. Recent progress in mechanistic interpretability techniques like activation patching~\citep{nanda2023attribution,heimersheim2024use,zhang2024towards}, path patching~\citep{goldowsky2023localizing}, logit lens analysis~\citep{wang2025logitlens4llms}, circuit tracing, and attribution graphs~\citep{ameisen2025circuit,lindsey2025biology} have been used to map specific behaviors (\eg subject–verb agreement, factual recall) to distributed circuits within LLMs. These tools allow researchers to \textbf{move beyond saliency} and identify what specific neurons or attention heads represent, and how information flows through a computation to produce an output. Yet, these techniques remain largely confined to unimodal models, particularly language models, and are computationally very expensive to interpret frontier models, limiting their research to industrial research labs~\citep{goh2021multimodal} and offering little to no reproducibility of research~\citep{circuit-tracer}. We lack analogous tools for multimodal systems and understanding of how modality-specific representations evolve or how signals from one modality are routed to influence representations in another (see Figure~\ref{fig:logit}). In particular, i) Multimodal models typically involve \textbf{heterogeneous} encoder stacks (\eg a ResNet for vision, a Graph Neural Network for graphs, and a transformer for text) with distinct representations and inductive biases, making it difficult to define shared ``\textit{units}'' (like neurons or heads) across modalities. ii) Fusion layers in multimodal models often entangle information from different modalities in ways that defy clean causal intervention: altering one modality's activation may have unpredictable effects on the joint representation. iii) There is no analog to ``\textit{circuits}'' in multimodal models, where cross-modal influence is often mediated through distributed, non-modular pathways. As a result, the methodological assumptions underlying mechanistic interpretability — \eg functionally meaningful subcircuits can be isolated and patched — do not yet transfer to these architectures, leaving a significant interpretability vacuum in understanding how and where multimodal reasoning occurs. 

% \subsection{The Limits of Sparse Autoencoders for Multimodal Interpretability}
Another prominent tool in mechanistic interpretability is Sparse autoencoders (SAEs) that have long been advocated in unimodal settings for learning compact, interpretable representations~\citep{cunningham2023sparse,makelov2024towards,bereska2024mechanistic}. Formally, SAEs are neural networks trained to reconstruct their input while enforcing a sparsity penalty on hidden activations~\citep{shu2025survey}, encouraging each input to be encoded using only a few ``\textit{active}'' neurons and learn a compact set of ``\textit{basis}'' features that can be individually inspected or sparsely activated to explain a prediction~\citep{makhzani2013k,tillsparse,o2024sparse,gao2024scaling,kissane2024sparse,kantamneni2025sparse,heap2025sparse}. However, to extend SAEs to multimodal models, one must do the following: a) train separate autoencoders per modality and concatenate their latent codes (losing sparsity guarantees), obscuring any shared structure or cross-modal interactions. Current SAEs \textbf{lack an explicit mechanism to capture cross-modal interactions}, \ie they treat each modality's features independently in their encoder, then fuse only at the latent layer, making it difficult to disentangle which sparse units represent joint versus modality-specific information, b) attempt to learn a joint autoencoder on the combined high-dimensional input (which quickly becomes intractable), c) scale sparsity constraints across heterogeneous data streams, requiring careful balancing of per-modality sparsity levels and aligning latent codes, which becomes non-trivial as modalities multiply and interact. In practice, SAEs fail to scale: they neither produce a unified, sparse latent space that reflects how modalities influence each other nor support direct interrogation of cross-modal reasoning. As a result, while they remain useful for unimodal interpretability, current forms of SAEs are ill-suited to explaining the synergistic behavior of multimodal models.

\looseness=-1 Finally, the shortcomings of unimodal explanations hinder our ability to \textbf{trust and diagnose multimodal models}. If the generated explanation cannot faithfully communicate how the model used individual modalities for its final decision, \textit{we risk false assurance and missed weaknesses}. For high-stakes applications, this is particularly problematic, \eg in an audio-text multimodal classifier, an explanation that only highlights words may not reveal that background noise in the audio was the true reason for the model's prediction~\citep{wu2023audio}. Without modality-aware explanations, we remain fundamentally uninformed about the internal mechanics of multimodal models, erroneously assuming that independent, per-modality attributions can be integrated post hoc into a coherent understanding -- they cannot. The next question is how to do better – \textit{What properties adequately illuminate a multimodal model's behavior?}

\section{Formal Desiderata for Multimodal Explanations}
\label{sec:multimodal}
\looseness=-1 Despite the growing deployment of multimodal models, current methods for explaining them are largely ad-hoc and poorly standardized. While certain techniques exist to explain individual modalities, these approaches do not accurately account for the interactions between modalities that are often crucial for understanding a model's decision~\citep{lyu2022dime,wenderoth2025measuring}. For example, in a vision-language model, the importance of an image region might change depending on the accompanying text, yet existing methods fail to model this interaction systematically. We argue that multimodal explanations must satisfy a set of core desiderata to move towards systematic multimodal explainability. These desiderata serve as both evaluation criteria for comparing explanation methods and design guidelines for developing new ones. Below, we first provide the relevant notations for the multimodal model.
%  and then discuss each desideratum and explain its significance.

\looseness=-1\xhdr{Notations} Let's consider a multimodal model $f : \mathcal{X}_1 \times \mathcal{X}_2 \times  \dots \mathcal{X}_M \to \mathcal{Y}$, where $\mathcal{X}_m$ is the input space for modality $m$, and  $M$ is the number of modalities model $f$ is trained on. Here, $\mathcal{Y}$ is not limited to just discrete decisions, but can comprise textual outputs or generated images as well. Let the input $\mathbf{x} = (x_1, x_2, \dots, x_M)$ denote a multimodal input, with $x_m \in \mathcal{X}_m$ the data for modality $m$, with prediction $f(\mathbf{x}) \in \mathcal{Y}$. We write $\mathbf{x}_{-m} = (x_1, \dots, x_{m-1}, x_m^{\mathrm{ref}}, x_{m+1}, \dots, x_M)$ for the same input but with modality $m$ replaced by a neutral reference $x_m^{\mathrm{ref}}$ (\eg an all-zero tensor). Further, let's consider an explanation function: $\mathcal{E} = \{\mathcal{E}_1, \mathcal{E}_2, \dots \mathcal{E}_M\}$, where each $\mathcal{E}_m: \mathcal{X}_m \times f \to \mathcal{Z}_m$ is an explanation object depending on the explanation type because we can have \textbf{homogeneous} explanations (same explanation method across modalities like gradient-based) or \textbf{heterogeneous} methods (different explanation methods per modality). For instance, $\mathcal{Z}_m \in \mathbb{R}^{d_m}$ could be feature-level attributions, $\mathcal{Z}_m \in \textrm{Concept Scores} \subseteq \mathbb{R}^k$ for concept-based methods, or $\mathcal{Z}_m \in $ Pathways or Circuits for mechanistic interpretability. Hence, the final composite explanation becomes:
% \abovedisplayshortskip=0pt
\begin{equation}
    \mathcal{E}(\mathbf{x}; f) = \bigcup_{m=1}^{M} \mathcal{E}_m(\mathbf{x}_m; f)
\end{equation}
% \belowdisplayshortskip=0pt
\looseness=-1 Note that the above explanation object also raises compatibility and unification challenges about heterogeneous explanation techniques across modalities, where some modalities (\eg text) are amenable to mechanistic interpretability (like attention head tracing, circuit analysis~\citep{ameisen2025circuit}) and others (\eg vision and audio) might lend themselves better to concept-based explanations (like TCAV~\citep{kim2018interpretability}, Concept Bottleneck Models~\citep{koh2020concept}) and some (\eg tabular data and graphs) might be best explained with feature attribution.

\begin{tcolorbox}[mybox]
A multimodal explanation is a structured interpretation of how multiple modalities jointly influence a model's prediction, reflecting both intra-modality reasoning and inter-modality interaction.
\end{tcolorbox}
% Thus, explanations must reflect both intra-modality reasoning and inter-modality interaction, all while satisfying the following properties.

\looseness=-1\xhdr{I) Granger Modality Influence} A multimodal explanation method should not merely correlate features with outputs but must reflect the model's predictive dependence on each input modality. For each modality $m$, we can define the model's ablated output change as: % , in the spirit of Granger causality
% \abovedisplayshortskip=0pt
\begin{equation}
    \Delta {f_m} = | f(\mathbf{x}) - f(\mathbf{x}_{-m}) |,
\end{equation}
% \belowdisplayshortskip=0pt
where $\mathbf{x}_{-m}$ replaces $\mathbf{x}_m$ with a neutral reference (\eg a blank image or empty text). A large $\Delta {f_m}$ indicates the model truly relies on modality $m$ to make its prediction. Let the explainer for modality  $m$ produce $\mathcal{E}_m(\mathbf{x})$ in some space $\mathcal{Z}_m$, and measure the explanation change as: $\Delta {E_m}(\mathbf{x}) = | \mathcal{E}_m(\mathbf{x}) - \mathcal{E}_{m}(\mathbf{x}_{-m}) |_{\mathcal{Z}_m},$ where $\mathcal{E}_m (\cdot)$ is the modality-specific explanation (attribution vector, concept scores, circuit activations, etc.) and $\| \cdot \|\mathcal{Z}_m$ is a suitable norm or distance in its explanation space.

We require that there exist constants $\alpha_m > 0$ and $\beta_m \geq 0$ such that, for all inputs: $\Delta {E_m}(\mathbf{x}) \geq \alpha_m \Delta {f_m} - \beta_m$, \ie if ablating modality $m$ changes the model output by $\Delta f_m$, the explainer must adjust its importance signal for $m$ by at least $\alpha_m \Delta {f_m} - \beta_m$. This ensures that the explainer's output for each modality is causally sensitive -- it reflects exactly how much the model truly relied on that modality, and flags any mismatch between the model's Granger-style dependence and the explanation's signal. It prevents an explainer from overstating or understating a modality's role. For example, if the model’s prediction collapses without the image ($\Delta f_{\mathrm{image}}$), yet $\mathcal{E}$ is near zero, the explanation is hiding a critical dependency.  If the model exploits a shortcut, then ablating the image yields $\Delta f_{\mathrm{image}} \approx 0$ and a reliable multimodal explainer must also show $\Delta E_{\mathrm{image}} \approx 0$, signaling that image cues were indeed not used.

\xhdr{II) Faithfulness} At its core, faithfulness demands that an explanation $\mathcal{E}$ reflect the actual computational reasoning of the model $f$. It is a cornerstone of any explanation and argues that if an explanation highlights certain input features, concepts, or neurons as important, then varying those factors should change the model's output accordingly. In a purely unimodal setting, we typically require that the explanation's importance scores align with how the model's output changes when we perturb or remove important features/concepts/circuits. Formally, an explanation $\mathcal{E}(\mathbf{x})$ is faithful if $\|f(\mathbf{x}) - f'(\mathbf{x})\| \approx 0$, where $f'(\mathbf{x})$ is the output of the model, where the irrelevant input features, concept activations, or intermediate neuron circuits are masked or zeroed out. %_{\mathrm{ablated}}

\begin{figure}[t]
    \centering
    \includegraphics[width=\textwidth]{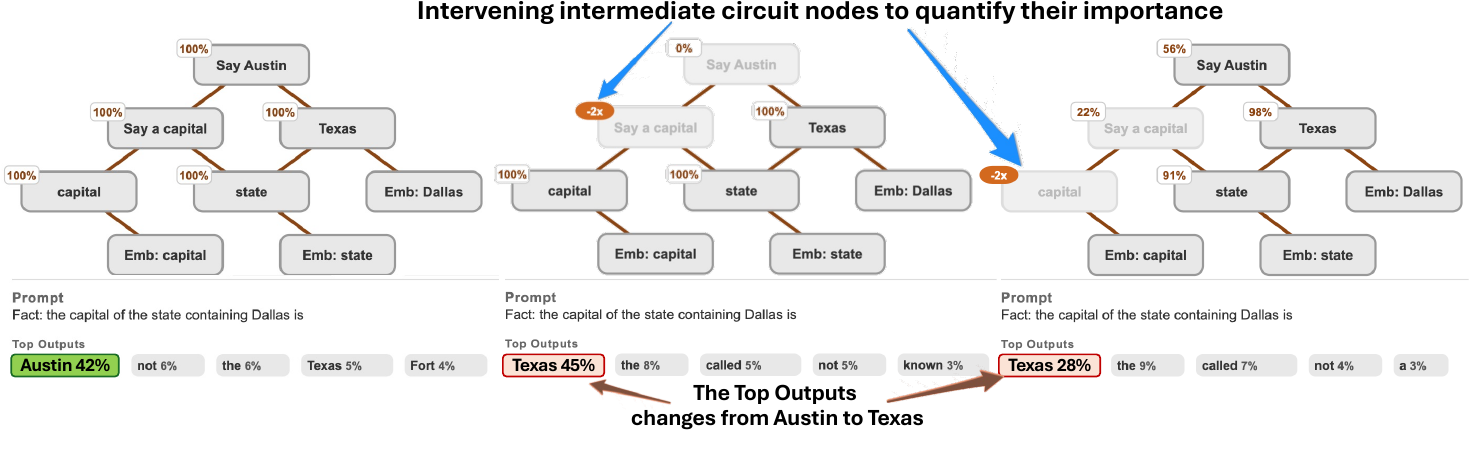}
    \vskip -0.15in
    \caption{\xhdr{Tracing and Quantifying Node Importance via Circuit Interventions in LLMs~\citep{circuit-tracer}} The figure illustrates the method of circuit tracing to identify and quantify the importance of intermediate nodes in a transformer-based LLM's reasoning process. The task involves answering the prompt: ``\textit{Fact: the capital of the state containing Dallas is}'', with the model initially outputting "Austin" (42\%) as the top prediction (\textbf{left panel}). Using~\citep{circuit-tracer}, the model's (Gemma-2-2b) reasoning is visualized as a compositional graph of nodes (\eg `Say Austin', `Texas', `capital') that derive their activation through embeddings (\eg \textit{Emb: capital, Emb: state}) and transformations. Each node is annotated with a contribution score (\eg 100\%, 56\%) representing its role in the original output. The \textbf{center} and \textbf{right} panels demonstrate the effect of intervening (ablating or suppressing) specific nodes by multiplying their activations by $-2\times$, thereby disrupting the information flow. Intervening on the ``\textit{Say a capital}'' node (\textbf{middle} panel) or the ``\textit{capital}'' node (\textbf{right} panel) causes the output to shift significantly from \textcolor{darkgreen}{\textbf{Austin}} to \textcolor{red}{\textbf{Texas}} (45\% and 28\% respectively), suggesting that these nodes play a critical role in enabling the \textbf{faithful} reasoning pathway.}
    \label{fig:circuit}
\end{figure}
In multimodal settings, each input modality (\eg image, text, audio) can contribute uniquely to a model's decision. A faithful explainer must identify exactly which parts of each modality the model uses. \textbf{Synergistic faithfulness} ensures this by using the explainer's own output to test sufficiency (``\textit{keep only what the explainer highlights}'') and necessity (``\textit{remove only what the explainer highlights}''). If the model's prediction remains unchanged when the highlighted parts remain and decreases when only those parts are removed, we gain confidence that the explainer has correctly identified the true marginal contribution of that modality. Synergistic faithfulness requires that whatever form the per-modality explanation $\mathcal{E}_m(\mathbf{x})$ takes (attributions, concept activations, attention patterns, or circuit traces), it can be used to construct two different feature/concept/circuit attributions whose effect on the model's output tests both sufficiency and necessity of the explained information. As described in the notations, for each modality $m$, the explainer produces an object $\mathcal{E}_m(\mathbf{x}) \in \mathcal{Z}_m$. For instance, in the case of feature attributions, the \textit{perturbation constructors} are:
% \abovedisplayshortskip=0pt
\begin{equation}
    \textrm{Keep}_{m} : \mathcal{Z}_{m} \times \mathcal{X}_{m} \to \mathcal{X}_{m}~~;~~  \textrm{Remove}_{m} : \mathcal{Z}_{m} \times \mathcal{X}_{m} \to \mathcal{X}_{m},
\end{equation}
% \belowdisplayshortskip=0pt
which takes the explanation and the original modality input $\mathbf{x}_m$ and return: i) $\textrm{Keep}_{m}(\mathcal{E}_{m}(\mathbf{x}), \mathbf{x}_{m})$, an input that \textit{retains} only the features (pixels, tokens, node and edge features, etc.) identified important by $\mathcal{E}_{m}$ and ii) $\textrm{Remove}_{m}(\mathcal{E}_{m}(\mathbf{x}), \mathbf{x}_{m})$ an input that precisely \textit{removes} the important features identified by $\mathcal{E}_{m}$. Next, we define the two perturbed full‐inputs: $\mathbf{x}^{\mathrm{keep}}_{m}{=}(\mathbf{x}_1, \dots, \mathbf{x}_{m-1}, \textrm{Keep}_{m}(\mathcal{E}_{m}(\mathbf{x}), \mathbf{x}_{m}), \mathbf{x}_{m+1}, \dots, \mathbf{x}_{M})$ and $\mathbf{x}^{\mathrm{remove}}_{m} = (\mathbf{x}_1, \dots, \mathbf{x}_{m-1}, \textrm{Remove}_{m}(\mathcal{E}_{m}(\mathbf{x}), \mathbf{x}_{m}), \mathbf{x}_{m+1}, \dots, \mathbf{x}_{M})$. Following these two perturbed inputs, we now define two different tests on sufficiency (keep) and necessity (remove), \ie

\xhdr{i) Sufficiency (keep)} The model's output on the input (identified as important by the underlying multimodal explainer) remains close to the original output:
\begin{equation}
    |~f(\mathbf{x}) - f(\mathbf{x}^{\mathrm{keep}}_{m})~| \leq \epsilon^{\mathrm{keep}}_{m},
    \label{eq:sufficiency}
\end{equation}
which ensures the input feature the explainer preserves indeed \textit{suffice} for the model's output decision.

\xhdr{ii) Necessity (remove)} The model's output on the removed input degrades significantly:
\begin{equation}
    |~f(\mathbf{x}) - f(\mathbf{x}^{\mathrm{remove}}_{m})~| \geq \epsilon^{\mathrm{remove}}_{m},
    \label{eq:necessity}
\end{equation}
which ensures the explainer's identified components are \textit{necessary} to the model's prediction.

It is to be noted that since $\mathcal{E}_m(\mathbf{x})$ can be any structured explanation, the perturbation constructors $\textrm{Keep}_{m}$ and $\textrm{Remove}_{m}$ must be implemented in a modality- and explainer-specific way (\eg \textbf{masking out pixels or tokens with low attribution, retaining only top-activated concepts, zeroing out non-critical circuit activations}). For instance, in mechanistic interpretability, one way to perform the above ablations is by intervening in the identified important circuit trace (see Fig.~\ref{fig:circuit}). A multimodal explainer passing both Eq.~\ref{eq:sufficiency} and Eq.~\ref{eq:necessity} for every modality $m$ constitutes \textit{rigorous synergistic faithfulness}, \ie the explanation itself drives exactly which parts of each modality to test, and the model's responses certify that those parts truly underlie its reasoning.

\looseness=-1\xhdr{III) Unified Stability} Unimodal explanation techniques are known to face stability issues, \ie small changes in input can lead to different explanations~\citep{alvarez2018robustness,agarwal2022rethinking}, which is exacerbated in multimodal settings. Stable explanations require accounting for how modalities trade off and complement each other. Moreover, when multimodal explanation methods generate modality-specific explanations (one per modality), their results can conflict or overlap arbitrarily, reducing overall consistency. Further, in multimodal explanation, stability issues can arise from robustness to modality variations (inter-modal stability) and semantic alignment (intra-modal stability). In particular, an explanation method 
$\mathcal{E} = \{\mathcal{E}_1, \mathcal{E}_2, \dots \mathcal{E}_M\}$ satisfies \textit{Unified Stability} if there exist tolerances $\{\epsilon_{m\to n}, \eta_m\}$ and Lipschitz constants $\{L_{n, m}, L'_m\}$ that satisfies the following bounds.

\xhdr{i) Modal-based Semantic Stability} For each modality $m$, and any pair of semantically equivalent inputs $\mathbf{x}_m$, $\mathbf{x}'_m$, \eg synonyms, perceptually invariant augmentations, satisfying $\| \mathbf{x}_m - \mathbf{x}'_m\| \leq \eta_m$:
\begin{equation}
    \| \mathcal{E}_m (\mathbf{x}_1, \dots, \mathbf{x}_m, \dots, \mathbf{x}_M) - \mathcal{E}_m (\mathbf{x}_1, \dots, \mathbf{x}'_m, \dots, \mathbf{x}_M) \| \leq L'_m \| \mathbf{x}_m - \mathbf{x}'_m \|,
    \label{eq:stability}
\end{equation}
\looseness=-1 where $L'_m$ is the Lipschitz constant, and the interpretation of the above property is that explanations for modality $m$ vary smoothly under \textit{semantic-preserving} changes to $\mathbf{x}_m$. Note that in Eqn.~\ref{eq:stability} the explanation method $\mathcal{E}_m$ takes the form of an attribution-based explanation that generates an importance score for each input feature. However, this can be easily extended to other explanation methods (like concept scores or internal circuits), and the Euclidean norm will be replaced by a corresponding distance metric.

\begin{figure}[t]
    \centering
    \includegraphics[width=0.95\textwidth]{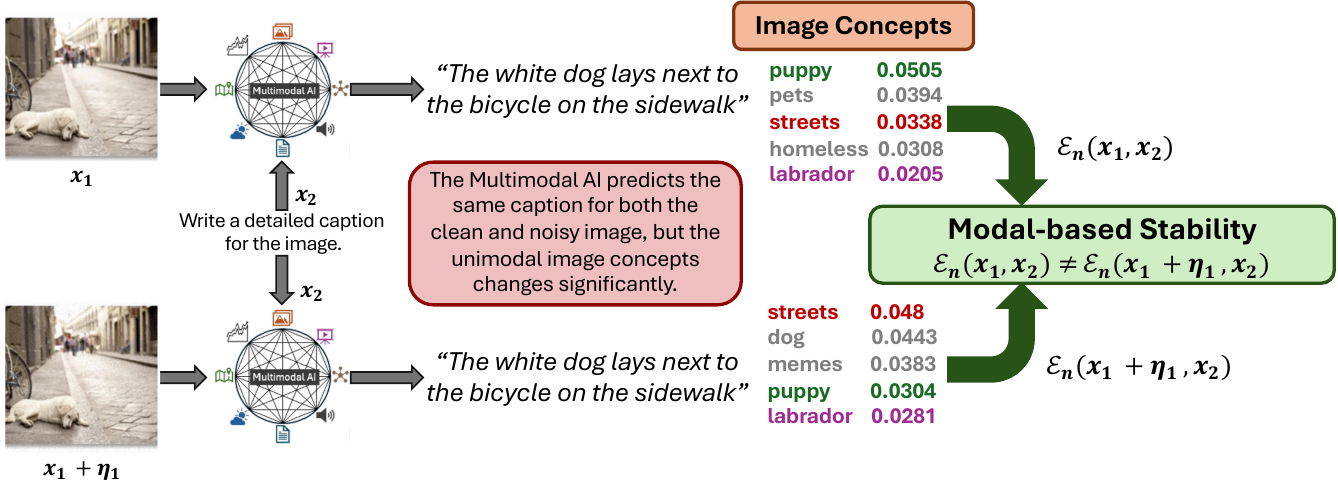}
    \caption{\xhdr{Evaluating Modal-based Stability with Concept-Based Explanations} Given a multimodal captioning model and an image $\mathbf{x}_1$ paired with text $\mathbf{x}_2$, we test cross-modal stability by applying a small, output-invariant perturbation $\delta_1$ to the image. While both the original $\mathbf{x}_1$ and perturbed $\mathbf{x}_2$ image yield the same caption: ``\textit{The white dog lays next to the bicycle on the sidewalk}'', the ranked concept scores differ significantly, revealing that small, semantically irrelevant image perturbations can produce large shifts in the explanation, undermining trust in the unimodal concept-based explanations. Notably, we observe spurious concepts like \textcolor{gray}{\textbf{memes}} getting a higher concept score than \textcolor{darkgreen}{\textbf{puppy}} and \textcolor{violet}{\textbf{labrador}} for the perturbed image.}
    \label{fig:stability}
\end{figure}
\xhdr{ii) Cross-Modal Stability} For any two distinct modalities $m \neq n$, any infinitesimal perturbation $\|\eta_m\| \leq \epsilon_{m\to n}$ in modality $m$ should have a minimal impact on the explanation for modality $n$, \ie
% \abovedisplayshortskip=0pt
\begin{equation}
    \| \mathcal{E}_n (\mathbf{x}_1, \dots, \mathbf{x}_m + \eta_m, \dots, \mathbf{x}_M) - \mathcal{E}_n (\mathbf{x}) \| \leq L_{n, m} \| \eta_m \|,
    \label{eq:crossmodal}
\end{equation}
% \belowdisplayshortskip=0pt
where $L_{n, m}$ is the Lipschitz constant and the interpretation is that small changes in one modality $m$ induce only small changes in the explanation for any other modality $n$. As before, the explanation $\mathcal{E}_n$ can take different forms. While cross-modal stability (Eqn.~\ref{eq:crossmodal}) ensures explanations for modality $n$ do not overreact when another modality $m$ is perturbed, capturing the original ``\textit{robustness to modality variations},'' intra-modal semantic consistency ensures explanations for modality $m$ respect meaning by capturing the original semantic alignment. By combining these into a \textbf{unified stability} constraint, any multimodal explanation method (whether mixing attribution, attention, concept, or mechanistic tools) must guarantee a bounded response to both kinds of permissible input changes (see Fig.~\ref{fig:stability}, where we generate concept explanations using Splice~\citep{bhalla2024interpreting}).

The above-mentioned desiderata (modality influence, synergistic faithfulness, and unified stability) form the core of what we propose as a framework for multimodal explainability, where they form a structured way to ensure that the explanations generated for multimodal models are reliable and transparently reflect the behavior of the underlying model.

\section{Conclusion and Call to Action}
\label{sec:conclusion}
\looseness=-1 While multimodal models are now at the forefront of AI breakthroughs, from multimodal dialogue agents to medical diagnosis systems that integrate diverse data, our ability to explain these models is lagging. The common practice of repurposing unimodal explainability techniques for multimodal models is fundamentally insufficient and potentially misleading. The explanations we get from such an approach are fragmented, lacking in representation of cross-modal interactions, and often unfaithful to what the model is truly doing. In light of this, our call to action is simple: the community must not naively rely only on unimodal explanation tools, and instead invest in developing modality-aware, behavior-grounded explainability methods tailored to multimodal reasoning.

\looseness=-1 \textit{\textbf{What might this entail?}} First, researchers should design new XAI methods that natively handle multiple modalities and produce a coherent explanation for a model's decision, encompassing all relevant modalities and their relationships. This could mean extensions of attribution techniques that explicitly compute interaction effects, visualization methods that jointly display multiple data streams with the model's focus, or even generating textual explanations that refer to multimodal evidence (\eg ``\textit{because the image shows X and the text says Y}''). Early work in this direction~\citep{lyu2022dime,wenderoth2025measuring,wang2021m2lens}
% (as cited in our discussion, \eg DIME, InterSHAP, and M2Lens) 
provides valuable proof-of-concept that multimodal explanations are feasible and can reveal insights hidden to unimodal analysis. We encourage more such innovation, particularly to handle complex architectures like transformers that blend modalities deeply. 

Second, we urge the development of evaluation frameworks and benchmarks for multimodal explainability. Just as the XAI field has devised ways to quantify explanation quality (faithfulness, fidelity, etc.) for unimodal models, we need analogous metrics and benchmark tasks for multimodal explanations. For instance, one could create synthetic multimodal tasks where the ground-truth contribution of each modality is known, to test whether an explanation method can recover it. Without rigorous evaluation, it will be hard to compare approaches or trust them in practice. The community should also gather feedback from users (\eg clinicians, users of multimedia systems) on what forms of multimodal explanations are most understandable and useful. Ensuring that modality-aware explanations are not only technically correct but also usable is important for their adoption. 

\looseness=-1 Third, there is a need for best practices and integration of explainability in the development of multimodal models themselves. If explainability is considered from the start, developers might choose model architectures that are more interpretable (for instance, using explicit alignment modules that can be visualized, or disentangled representations that separate modality contributions). At the very least, practitioners deploying multimodal models should be cautious of relying on off-the-shelf unimodal explanation tools. We recommend treating any such explanation with skepticism – if one must use them, combine them and look for consistency (\textit{does the text explanation align with the image explanation?}), and recognize what they might be missing (interactions, hidden biases). 
% The ultimate solution, however, is to have explanation tools purpose-built for these models, so that such ad-hoc judgments are not needed. 

\looseness=-1 In conclusion, unimodal explanations for multimodal models are a misfit, \ie they fail to explain what is truly a joint, multimodal decision process. The field of XAI must evolve in step with the models it aims to explain. By developing multimodal explainability techniques that capture attribution across modalities, highlight cross-modal interactions, align with model internals, and maintain faithfulness, we can ensure that our understanding of AI keeps up with its capabilities. This is not only an intellectual necessity but also a practical one: as AI systems integrate into critical domains (healthcare, law, transportation) using rich multimodal data, our ability to interpret and trust these systems will hinge on the quality of explanations they can provide. We contend that the current maturity of multimodal AI warrants a dedicated research focused on the development of \textbf{modality-aware, behavior-grounded explanation} algorithms.
% We believe the time is perfect for a concerted effort on developing \textbf{modality-aware, behavior-grounded explanation} algorithms. 
The insights and examples discussed in this paper serve as a conceptual foundation – it is time to stop using unimodal explainability for multimodal models, and start building explanation tools that the multimodal age demands.

\begin{ack}
The author would like to thank Guadalupe González, Suraj Srinivas, Anh Nguyen, Soham Petkar, and all members of \href{https://chirag-agarwall.github.io/}{Aikyam Lab} for their insightful discussions and valuable feedback. This work is supported in part by research awards from LaCross AI Institute and Cohere. CA would like to thank Nikita Kharya for her continued support and encouragement. The views expressed here are
those of the authors and do not reflect the official policy or position of the funding agencies.
\end{ack}

\newpage
\bibliographystyle{plainnat}
\bibliography{references}

% \section*{References}
% References follow the acknowledgments in the camera-ready paper. Use unnumbered first-level heading for
% the references. Any choice of citation style is acceptable as long as you are
% consistent. It is permissible to reduce the font size to \verb+small+ (9 point)
% when listing the references.
% Note that the Reference section does not count towards the page limit.
\medskip

% {
% \small

% [1] Alexander, J.A.\ \& Mozer, M.C.\ (1995) Template-based algorithms for
% connectionist rule extraction. In G.\ Tesauro, D.S.\ Touretzky and T.K.\ Leen
% (eds.), {\it Advances in Neural Information Processing Systems 7},
% pp.\ 609--616. Cambridge, MA: MIT Press.

% [2] Bower, J.M.\ \& Beeman, D.\ (1995) {\it The Book of GENESIS: Exploring
%   Realistic Neural Models with the GEneral NEural SImulation System.}  New York:
% TELOS/Springer--Verlag.

% [3] Hasselmo, M.E., Schnell, E.\ \& Barkai, E.\ (1995) Dynamics of learning and
% recall at excitatory recurrent synapses and cholinergic modulation in rat
% hippocampal region CA3. {\it Journal of Neuroscience} {\bf 15}(7):5249-5262.
% }

%%%%%%%%%%%%%%%%%%%%%%%%%%%%%%%%%%%%%%%%%%%%%%%%%%%%%%%%%%%%
% \newpage
\appendix
\section{Appendix}
\label{app:experiment}

\xhdr{Experimental details for Logit Lens} For our experimental evaluation, we employed the Token Embedding-Aligned Graph Language Model (TEA-GLM)~\citep{wang2024llms} to perform zero-shot graph learning on the Cora citation network dataset~\citep{T8/HUIG48_2017}. This benchmark dataset is widely recognized in the graph learning community for node classification tasks.

The graph neural network (GNN) component was pretrained to align its representations with the token embeddings of an instruction-fine-tuned Phi-3.5 7B large language model~\citep{abdin2024phi}. Following this alignment process, we train a linear projector to map GNN representations to a fixed set of graph token embeddings specifically for the node classification task. All experiments were conducted on an NVIDIA RTX 5000 Ada Generation Graphics Card with 32GB of memory.

\begin{figure*}
    \centering
    \includegraphics[width=\textwidth, angle=90, scale=1.5]{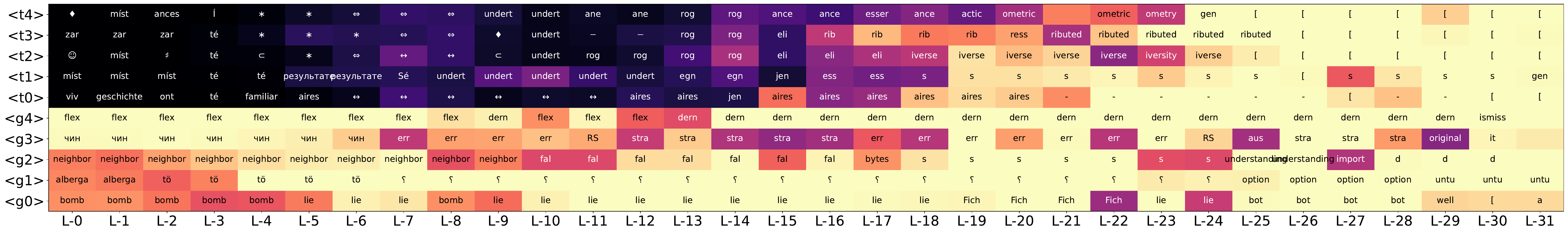}
    \caption{\xhdr{Interpreting Graph Language Models using Logit Lens}}
    \label{app:logit}
\end{figure*}

To investigate the interpretability challenges, we implemented the logit lens technique to analyze the model's predictive behavior on graph-related tasks (see Fig.~\ref{app:logit}). This method extracts token-level predictions at each layer by applying a softmax operation to the product of the language model's output layer weights (lm\_head) and the hidden state representations at each layer. This procedure is designed to reveal the model's ``thinking process'' by exposing what token the model would predict at each intermediate layer, providing insight into how representations evolve throughout the network. Our findings demonstrate that while this technique provides meaningful insights for text tokens, graph token representations exhibit minimal evolution across layers, confirming the inapplicability of this unimodal XAI method to multimodal architectures.

\end{document}